%% file: main.tex
\def\arxivmode{1}
\documentclass[10pt]{article}
\ifdefined\arxivmode\usepackage[preprint]{tmlr}\else\usepackage{tmlr}\fi

\usepackage{amsmath,amssymb}
\usepackage{booktabs}
\usepackage{longtable}
\usepackage{graphicx}
\usepackage{enumitem}
\usepackage{hyperref}
\usepackage{url}
\usepackage{acro}
\graphicspath{{figures/}}
\input{generated/numbers.tex}   % GENERATED inline-number macros (paper_v5/tools/build_tables.py)
\input{generated/glossary.tex}  % GENERATED vocabulary: \ac{TERM}, \arm{x} (paper_v5/tools/build_prose.py)
\input{generated/equations.tex} % GENERATED equations as macros: \eqDs ... (paper_v5/tools/build_prose.py)

\title{A JEPA Recipe for Tabular Foundation Models}

\author{\name Mingyu Jeon \email jkmcoma7@gmail.com \\
      \addr Modulabs, Seoul, Republic of Korea
      \AND
      \name Suwan Cho \email cho.suwan96@gmail.com \\
      \addr Modulabs, Seoul, Republic of Korea
      \AND
      \name Jae Young Suh \email tjwodud04@gmail.com \\
      \addr Modulabs, Seoul, Republic of Korea}

\def\month{09}
\def\year{2026}
\def\openreview{\url{https://openreview.net/forum?id=XXXX}}

\begin{document}
\maketitle

\input{sections/abstract.tex}
\input{sections/intro.tex}
\input{sections/related.tex}
\input{sections/method.tex}
\input{sections/experiments.tex}
\input{sections/discussion.tex}
\input{sections/conclusion.tex}

\bibliography{references}
\bibliographystyle{tmlr}

\appendix
\input{sections/appendix.tex}

\end{document}

%% file: generated/numbers.tex
\newcommand{\numBenchArmGapMax}{.068}
\newcommand{\numBenchArmGapMaxClf}{.010}
\newcommand{\numCpEncClf}{.750}
\newcommand{\numCpEncReg}{.490}

\newcommand{\numCpPredClfHi}{.585}
\newcommand{\numCpPredClfLo}{.570}
\newcommand{\numCpPredConstGapPctMax}{1.3}
\newcommand{\numCpPredErankMax}{26.7}
\newcommand{\numCpPredRegMax}{-.003}
\newcommand{\numCpRuleBackFrac}{.98}
\newcommand{\numCpRulePredCosFlat}{.9}
\newcommand{\numCpRuleRank}{10}
\newcommand{\numCpScmBackStep}{3,250}
\newcommand{\numCpScmMinStep}{2,250}
\newcommand{\numCpScmPredFlatStep}{2,750}
\newcommand{\numCpScmRankStep}{5,500}
\newcommand{\numCpTabiclOneMinOverConst}{1.000}
\newcommand{\numCpTabiclOnePredErankMax}{15.4}
\newcommand{\numCpTabiclTwoBackStep}{12,250}
\newcommand{\numCpTabiclTwoMinStep}{7,250}
\newcommand{\numCpTabiclTwoPredFlatStep}{8,750}
\newcommand{\numCpTabiclTwoRankStep}{13,500}
\newcommand{\numCurriculumSteps}{4,000}
\newcommand{\numDedupBothRules}{2}

\newcommand{\numDedupDupNamesClf}{12}
\newcommand{\numDedupDupNamesReg}{2}
\newcommand{\numDedupNClf}{102}
\newcommand{\numDedupNReg}{30}
\newcommand{\numDedupPairAll}{29:63}
\newcommand{\numDedupPairAllP}{.00051}

\newcommand{\numDedupPairReg}{8:22}
\newcommand{\numDedupPairRegP}{.01612}
\newcommand{\numDedupRuleMaxShift}{1}
\newcommand{\numDsClfMean}{.815}
\newcommand{\numDsCrossClf}{60,000}
\newcommand{\numDsCrossReg}{20,000}
\newcommand{\numDsFinalValMse}{.4179}
\newcommand{\numDsHours}{24.8}

\newcommand{\numDsStopStep}{178,500}
\newcommand{\numDsValMseTwentyk}{.5041}
\newcommand{\numDsValTables}{32}
\newcommand{\numDsVsHistgbClf}{28:84}
\newcommand{\numDsVsHistgbReg}{4:28}
\newcommand{\numDsVsLogregClf}{71:39}
\newcommand{\numDsVsRidgeReg}{27:5}
\newcommand{\numEma}{.996}
\newcommand{\numEmb}{256}
\newcommand{\numEncFinalErank}{98.1}
\newcommand{\numEncFinalValMse}{.5513}
\newcommand{\numFloorClf}{51:49}
\newcommand{\numFloorReg}{17:15}
\newcommand{\numGrinsztajnTwiceNames}{9}
\newcommand{\numHeads}{8}
\newcommand{\numHistgbClfMean}{.842}

\newcommand{\numJepaBestStepRatio}{1.47}

\newcommand{\numJepaConstMap}{.949}
\newcommand{\numJepaCrossClf}{140,000}
\newcommand{\numJepaCrossReg}{60,000}
\newcommand{\numJepaErank}{104.5}
\newcommand{\numJepaFinalValMse}{.4360}
\newcommand{\numJepaFirstBelow}{250}
\newcommand{\numJepaHours}{41.3}
\newcommand{\numJepaHoursRatio}{1.66}
\newcommand{\numJepaLatentLastWindow}{+.0094}

\newcommand{\numJepaStepRatio}{1.42}
\newcommand{\numJepaStopStep}{253,750}
\newcommand{\numJepaTgtCos}{.4398}
\newcommand{\numJepaValJepa}{.2209}

\newcommand{\numJepaValTables}{64}
\newcommand{\numJepaVsHistgbClf}{27:87}
\newcommand{\numJepaVsHistgbReg}{3:29}
\newcommand{\numJepaVsLogregClf}{66:44}
\newcommand{\numJepaVsRidgeReg}{25:7}
\newcommand{\numLambdaJepa}{1.0}
\newcommand{\numLambdaPpd}{1.0}
\newcommand{\numLayers}{6}

\newcommand{\numLr}{$5\times 10^{-4}$}
\newcommand{\numMaxCells}{65,536}
\newcommand{\numMaxClasses}{10}
\newcommand{\numMaxFeatures}{100}

\newcommand{\numNBins}{32}
\newcommand{\numNClf}{115}

\newcommand{\numNReg}{32}
\newcommand{\numNSets}{147}
\newcommand{\numPCls}{.7}
\newcommand{\numPMissing}{.3}
\newcommand{\numPairAll}{32:70}
\newcommand{\numPairAllAuc}{31:82}
\newcommand{\numPairAllAucTies}{2}
\newcommand{\numPairAllGap}{.006}
\newcommand{\numPairAllMeanDs}{.815}
\newcommand{\numPairAllMeanJepa}{.809}
\newcommand{\numPairAllP}{.00021}
\newcommand{\numPairAllTies}{13}
\newcommand{\numPairBinary}{24:44}
\newcommand{\numPairBinaryGap}{.003}

\newcommand{\numPairBinaryP}{.02053}
\newcommand{\numPairBudget}{20,000}
\newcommand{\numPairFewclass}{6:11}

\newcommand{\numPairFewclassP}{.33231}

\newcommand{\numPairGrinsztajnReg}{6:15}
\newcommand{\numPairGrinsztajnRegP}{.07835}
\newcommand{\numPairManyclass}{2:15}
\newcommand{\numPairManyclassGap}{.017}

\newcommand{\numPairManyclassP}{.00235}
\newcommand{\numPairOpenmlClf}{13:42}
\newcommand{\numPairOpenmlClfP}{.00011}
\newcommand{\numPairReg}{8:24}
\newcommand{\numPairRegGap}{.030}

\newcommand{\numPairRegP}{.007}

\newcommand{\numPairTabarenaReg}{2:9}
\newcommand{\numPairTabarenaRegP}{.06543}
\newcommand{\numPatienceSteps}{20,000}
\newcommand{\numPredConstMap}{.949}
\newcommand{\numPredFinalErank}{1.5}
\newcommand{\numPredFinalLatentLoss}{.0083}
\newcommand{\numPredFinalValMse}{.9364}
\newcommand{\numPredMinStep}{7,250}
\newcommand{\numPredMinValMse}{.7499}
\newcommand{\numPredictorDepth}{4}
\newcommand{\numRegDsBelowRidge}{5}
\newcommand{\numRegGapMedian}{.011}
\newcommand{\numRegJepaBelowRidge}{7}
\newcommand{\numRegTopGapOne}{.347}
\newcommand{\numRegTopGapTwo}{.156}
\newcommand{\numRegTopNameOne}{concrete compressive strength}
\newcommand{\numRegTopNameTwo}{airfoil self noise}
\newcommand{\numRegTopTwoShareTabarena}{68}

\newcommand{\numRowsMax}{1,024}
\newcommand{\numRowsMin}{128}
\newcommand{\numSnapEvery}{5,000}
\newcommand{\numSnapScoreEvery}{20,000}
\newcommand{\numStepsCap}{1,000,000}
\newcommand{\numStopDelta}{.002}
\newcommand{\numStopPatience}{80}
\newcommand{\numSuiteCtx}{1,024}
\newcommand{\numSuiteKfeat}{64}
\newcommand{\numSuiteNSplits}{3}
\newcommand{\numSuiteTestCap}{512}
\newcommand{\numTrajFinalFirst}{9:96}
\newcommand{\numTrajFinalFirstStep}{20,000}
\newcommand{\numTrajFinalLast}{31:69}
\newcommand{\numTrajFinalLastStep}{240,000}
\newcommand{\numTrajSameFirst}{22:80}
\newcommand{\numTrajSameFirstStep}{20,000}
\newcommand{\numTrajSameLast}{17:79}
\newcommand{\numTrajSameLastStep}{160,000}
\newcommand{\numValEvery}{250}

%% file: generated/glossary.tex
\DeclareAcronym{ICL}{short=ICL, long={in-context learning}}
\DeclareAcronym{PFN}{short=PFN, long={prior-fitted network}}
\DeclareAcronym{JEPA}{short=JEPA, long={joint-embedding predictive architecture}}
\DeclareAcronym{EMA}{short=EMA, long={exponential moving average}}
\DeclareAcronym{SSL}{short=SSL, long={self-supervised learning}}
\DeclareAcronym{SCM}{short=SCM, long={structural causal model}}
\DeclareAcronym{MLP}{short=MLP, long={multi-layer perceptron}}
\newcommand{\arm}[1]{\csname arm@#1\endcsname}
\expandafter\def\csname arm@ds\endcsname{\textsc{ds}}
\expandafter\def\csname arm@jepa\endcsname{\textsc{jepa}}

%% file: generated/equations.tex
\newcommand{\eqDs}{%
\begin{equation}
  \mathcal{L}_{\mathrm{ds}}=\frac{1}{|T|}\sum_{(i,j)\in T}\mathrm{CE}\big(\psi(h_{ij}),\,b(x_{ij})\big).
  \label{eq:ds}
\end{equation}}
\newcommand{\eqJepa}{%
\begin{equation}
  \mathcal{L}_{\mathrm{jepa}}=\lambda_{\mathrm{jepa}}\,\frac{1}{d\,|M|}\sum_{(i,j)\in M}\big\lVert g(\bar h)_{ij}-\mathrm{sg}\big[\mathrm{LN}\big(\tilde z^{\mathrm{full}}_{ij}-\tilde z^{\mathrm{mask}}_{ij}\big)\big]\big\rVert_2^2\;+\;\lambda_{\mathrm{ds}}\,\mathcal{L}_{\mathrm{ds}}(h).
  \label{eq:jepa}
\end{equation}}

%% file: sections/abstract.tex
% GENERATED from paper_v5/{structure,claims,filler}.json by paper_v5/tools/build_prose.py -- do not edit.

\begin{abstract}
% ¶ abs
Tabular foundation models learn to predict cell values in context, whereas world-model self-supervision asks for prediction in representation space~\citep{lecun2022path,assran2023ijepa}.
On a tabular foundation-model prior, the latent term of a \ac{JEPA} collapsed in our earlier runs and took the encoder with it to a constant map.
We report a recipe under which the latent term survives to convergence beside the value objective: the value head reads the encoder field rather than the predictor, and the target is an \ac{EMA} difference.
To bound its cost against the value-only arm, both arms train until a plateau rule stops them, with no fixed step budget.
A fixed horizon had confounded a slowdown with a ceiling, since the value-only arm was still improving well past the usual budget.
At convergence, in one run per arm, the \ac{JEPA} arm trails the value-only arm across \numNSets{} real datasets, \numPairAll{} wins to losses on classification (\numDedupPairAll{} with one entry per dataset name) and \numPairReg{} on regression, the margin small on classification and wider on regression, and the count leans the same way in each stratum and each benchmark.
The \ac{JEPA} arm (\arm{jepa}) needs \numJepaStepRatio{} times as many steps as the value-only arm (\arm{ds}), and \numJepaHoursRatio{} times its wall-clock, to reach its plateau.

\end{abstract}
\acresetall  % the body gets its own first use of every term

%% file: sections/intro.tex
% GENERATED from paper_v5/{structure,claims,filler}.json by paper_v5/tools/build_prose.py -- do not edit.

\section{Introduction}
\label{sec:intro}

% ¶ intro-hook
Tabular foundation models such as TabPFN and TabICL learn in-context prediction from synthetic tables drawn from a prior, with no per-dataset training~\citep{hollmann2023tabpfn,hollmann2025tabpfnv2,qu2025tabicl}.
Their training objective predicts the value of a hidden cell from the visible rows, so the representation is shaped by a data-space loss alone~\citep{muller2021transformers}.
The world-model prescription for self-supervised learning is the opposite: predict the representation of the missing part, never its pixels or values~\citep{lecun2022path,assran2023ijepa}.
The two objectives pull the representation in different directions, since a value loss rewards whatever predicts the cell and a latent loss whatever predicts its representation~\citep{lecun2022path}, and on a synthetic prior nothing guarantees that the two agree.
A synthetic prior also makes such a comparison unusually clean, since the data are unlimited and drawn from one distribution for both objectives (\S\ref{sec:method}).
A difference between two arms is then a property of the objectives and their wiring rather than of a dataset (\S\ref{sec:method}).
Whether such a latent objective can be trained on a tabular foundation-model prior at all, and what it costs, is the question we answer.

% ¶ intro-obstacle
In our earlier runs a \ac{JEPA} latent term collapsed the encoder to a constant map on this prior, and the encoder ended at a constant map on every other prior we trained it on, on one of them never having left it (\S\ref{sec:method}, \S\ref{app:collapse}).
The collapse took the value head with it, because the head read a predictor output the latent term had already flattened.
The recipe below is what removed that failure.

% ¶ intro-contrib
We make three contributions.
First, a recipe that trains a \ac{JEPA} term beside the value objective on a tabular foundation-model prior without collapse: the value head reads the encoder field, the latent target is an \ac{EMA} difference, and hidden cells enter the predictor as mask tokens, of which the runs isolate the first, with the other two held fixed (\S\ref{sec:method}).
Second, an open-horizon protocol that stops each arm at a plateau, which showed that the fixed budget of our earlier comparisons had stopped the value-only arm short of its plateau (\S\ref{sec:convergence}).
Third, a converged comparison against the value-only arm on \numNSets{} real datasets, stratified by number of classes, whose sign test carries a noise floor from a second run of the value-only arm (\S\ref{sec:setup}, \S\ref{sec:final}).

% ¶ intro-scope
The recipe does not beat the value-only arm at convergence, and we do not claim that it does (\S\ref{sec:final}).
The claim is narrower, that the latent term trains on this prior (\S\ref{sec:convergence}) and that what the recipe costs at convergence is bounded from above by one run per arm, on the premise that no other difference between the arms helps the one with the term (\S\ref{sec:final}).

%% file: sections/related.tex
% GENERATED from paper_v5/{structure,claims,filler}.json by paper_v5/tools/build_prose.py -- do not edit.

\section{Related work}
\label{sec:related}

% ¶ rel-tabular-fms
\Acp{PFN} cast tabular prediction as amortised Bayesian inference over a synthetic prior, and TabPFN and TabICL scaled the idea to thousands of context rows~\citep{muller2021transformers,hollmann2025tabpfnv2,qu2025tabicl}.
Follow-up work attributes much of the family's behaviour to the prior rather than to the architecture~\citep{bouadi2026shapingprior}.
The family has since been trained on real tables, with in-context retrieval and self-supervision~\citep{ma2025tabdpt}, and on curated mixtures of synthetic priors~\citep{zhang2025mitra}.
Against that line, we keep the family's architecture, prior and value objective fixed and add one latent term, so that the comparison bounds what that term costs (\S\ref{sec:threats}).

% ¶ rel-jepa
A \ac{JEPA} predicts the representation of masked content from the visible context through a predictor, and avoids collapse with an \ac{EMA} target encoder or an explicit regulariser~\citep{assran2023ijepa,grill2020byol,bardes2022vicreg,balestriero2025lejepa}.
Its ancestors avoided collapse by a momentum target~\citep{grill2020byol}, self-distillation from a masked view~\citep{baevski2022data2vec}, a stop-gradient~\citep{chen2020simsiam} or a redundancy penalty~\citep{zbontar2021barlow,bardes2022vicreg}.
On video, feature prediction alone, with no reconstruction and no pretrained encoder, has since been shown to carry representation learning~\citep{bardes2024vjepa}.
Collapse to a constant or, in its partial form, to a low-dimensional subspace is the documented failure of that line~\citep{jing2021dimensional}, tracked by rank-based diagnostics~\citep{garrido2023rankme,thilak2023lidar,littwin2024jepabias}.
This paper reports that same failure on a tabular foundation-model prior and a wiring that removes it, rather than a new regulariser.

% ¶ rel-tabular-ssl
Tabular self-supervision has used corruption and reconstruction of rows~\citep{yoon2020vime,bahri2022scarf,ucar2021subtab,somepalli2021saint}, and \ac{JEPA} variants have been trained on single datasets~\citep{thimonier2024tjepa}.
Self-supervision has also come from few-shot tasks generated by treating columns of an unlabeled table as labels~\citep{nam2023stunt}, and pre-training has crossed tables without matched columns through a graph representation of table entries with string embeddings of entries and column names~\citep{kim2024carte}.
Cross-table pre-training transfers a backbone across tables whose columns differ, with self-supervised~\citep{zhu2023xtab} or mixed supervised and self-supervised objectives~\citep{wang2022transtab}.
LaT-PFN moved \ac{PFN} prediction into a latent space with a decoder~\citep{verdenius2024latpfn}.
Unlike those lines, we train the latent term on the prior itself, at \numRowsMin{} to \numRowsMax{} rows of context, and score the result on the family's real-data suite.

%% file: sections/method.tex
% GENERATED from paper_v5/{structure,claims,filler}.json by paper_v5/tools/build_prose.py -- do not edit.

\section{Method}
\label{sec:method}

\begin{figure}[t]
\centering
\includegraphics[width=\linewidth]{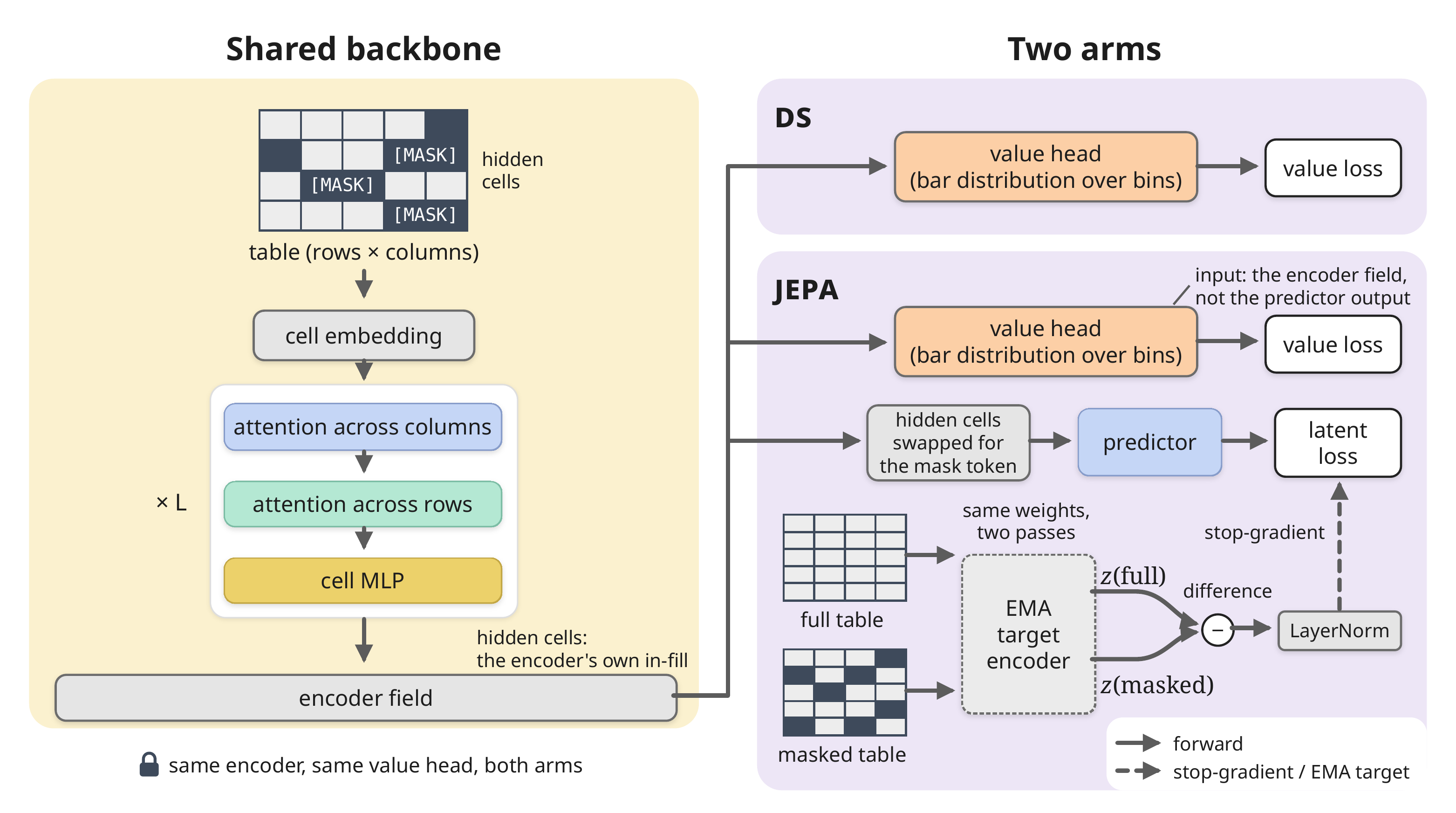}
\caption{\textbf{One backbone, two arms.} Both arms share the cell-level encoder and the value head that reads its field. The \arm{jepa} arm adds a predictor that reads the encoder field with hidden cells swapped for the mask token, and a latent loss against the stop-gradient difference of an \ac{EMA} target encoder's full-table and masked-table embeddings.}
\label{fig:arch}
\end{figure}

% ¶ method-backbone
We write \arm{ds} for the value-only arm and \arm{jepa} for the \ac{JEPA} arm.
Both arms share one cell-level transformer with \numLayers{} layers, \numHeads{} heads and width \numEmb{}, which encodes every cell of a table and reads the value of a hidden cell through a \ac{MLP} head over \numNBins{} bins.
The value objective is the cross-entropy of that head against the binned value of each hidden cell: \eqDs
\arm{ds} trains on this objective alone, with hidden cells drawn independently across the table.
Everything the two arms share is in this backbone and this head, and the recipe changes only what the head reads and what else the encoder is trained to predict (Figure~\ref{fig:arch}).

% ¶ method-recipe

\arm{jepa} adds a latent term and changes where the value head reads.
The value head reads the encoder field, where every hidden cell carries the encoder's own in-fill, instead of the predictor output that the latent term shapes (Figure~\ref{fig:arch}).
This wiring kept the latent term alive to the plateau, whereas the predictor-read head had collapsed early in our earlier runs (Figure~\ref{fig:curves}c).
That failure is shown by a pair of runs of the same recipe to the fixed budget of \numPairBudget{} steps, which differ in one configuration key, where the value head reads (\S\ref{app:config}).
With the head on the predictor the value error fell to \numPredMinValMse{} by step \numPredMinStep{} and rose back to \numPredFinalValMse{} by the budget, against a constant map of \numPredConstMap{}.
Target cosine is the mean pairwise cosine between the latent targets of the held-out cells and target effective rank the entropy rank of their \numEmb{}-dimensional matrix, whose ceiling is \numEmb{} and which a collapsed target drives toward one (Figure~\ref{fig:curves}c).
Its target effective rank ended at \numPredFinalErank{} and its latent loss at \numPredFinalLatentLoss{}, a target that has collapsed rather than been learned.
That wiring ended at the constant map of its own held-out set on the two other priors it was trained on as well (\S\ref{app:collapse}).
The same run with its head on the encoder field reached \numEncFinalValMse{} at the budget, with a target effective rank of \numEncFinalErank{} (Figure~\ref{fig:curves}c).
The latent target of a hidden cell is the \ac{EMA} target encoder's embedding of the full table minus its embedding of the masked table, layer-normalised and detached.
The difference carries what the hidden cells contributed to the field rather than a copy of the context, so that the predictor cannot satisfy the target from the visible cells alone.
The targets of data2vec and I-JEPA are the target encoder's representation of the full input at the hidden positions~\citep{baevski2022data2vec,assran2023ijepa}, made cheap to build by the successor of data2vec~\citep{baevski2022data2vec2}, and the difference target departs from them by subtracting the masked-table embedding.
The predictor, a \numPredictorDepth{}-block transformer, reads the encoder field with hidden cells replaced by a mask token, so that hidden cells are invisible to each other, as in the image \ac{JEPA} of \citet{assran2023ijepa}.
The context split is that second pass, in which the encoder of \arm{jepa} reads the visible cells only and the predictor fills the hidden ones from its mask token, while the value head reads the first pass over the masked table as in \arm{ds} (Figure~\ref{fig:arch}).
The objective of \arm{jepa} sums the latent loss, a mean over the $|M|$ hidden cells and the $d=\numEmb{}$ dimensions of their targets, and the value loss with weights \numLambdaJepa{} and \numLambdaPpd{}: \eqJepa
The prescription's own protocol is two-stage, a latent-only pretraining followed by a probe or a fine-tune on the task~\citep{assran2023ijepa,bardes2024vjepa}.
We test its objective in one stage beside the value loss instead, at one weight each, carried over rather than swept for this pair, since the family's task is value prediction on the same prior and there is no second task to adapt to.
The two-stage form needs a value read-out of its own, a probe on the frozen field, and is left to \S\ref{sec:future}.
Hidden cells are drawn under a mixed policy chosen per batch, as independent cells, as column blocks or as rectangular blocks, after a warm-up of \numCurriculumSteps{} steps on target-column masking.
The target encoder follows the online encoder with decay \numEma{}.

% ¶ method-prior
Both arms train on tables from the graph \ac{SCM} generator of TabICL~\citep{qu2025tabicl}, the second version of its prior, with \numRowsMin{} to \numRowsMax{} rows and up to \numMaxFeatures{} features per table.
The generator is the public TabICL code at commit 8f1aa20 of its repository, vendored with two edits, to its import paths and to one set-to-sorted call that had made the table stream differ between processes.
Both of its dataset filters are on, the one that rejects graphs whose feature nodes share no ancestor with the label and the one that rejects tables an extra-trees model cannot predict.
Each batch is a classification task with probability \numPCls{}, with up to \numMaxClasses{} classes, and a regression task otherwise, and a fraction \numPMissing{} of the tables carry missing feature cells.
Both dimensions sit under a budget of \numMaxCells{} cells per table, which bounds memory, so rows and features trade off within it.
Both arms draw the same stream of tables from the same seed, so the tables they train on are the same and differ only in how each arm splits and masks them (\S\ref{sec:discussion}).

% ¶ method-protocol
Instead of a fixed step budget, each arm trains until its best validation MSE has failed to improve by more than \numStopDelta{} for \numStopPatience{} consecutive validations, one every \numValEvery{} steps, under a safety cap of \numStepsCap{} steps.
The optimiser is schedule-free AdamW at learning rate \numLr{}, so that no learning-rate schedule is tied to the step cap.
The rule is the same for both arms and reads only the value error, so the arm with the latent term earns no budget beyond what its own value curve justifies (\S\ref{sec:convergence}).
Along that open horizon, a checkpoint is kept every \numSnapEvery{} steps and one every \numSnapScoreEvery{} steps is scored on the real-data suite, so that the comparison is available along the whole trajectory.

%% file: sections/experiments.tex
% GENERATED from paper_v5/{structure,claims,filler}.json by paper_v5/tools/build_prose.py -- do not edit.

\section{Experiments}
\label{sec:exp}

% ¶ exp-preamble
We compare the two arms at their plateaus and along the way.
The experiments ask whether the latent term trains to a plateau under the same stop rule as \arm{ds} (\S\ref{sec:convergence}), what the recipe costs once both arms have converged (\S\ref{sec:final}), and where each arm stands against classical baselines along the way (\S\ref{sec:baselines}).
Every number comes from one seed per arm, with the noise floor of the sign test taken from a repeat of \arm{ds} (\S\ref{sec:setup}).

\subsection{Setup}
\label{sec:setup}

% ¶ exp-setup
The real-data suite holds \numNClf{} classification and \numNReg{} regression datasets from OpenML-CC18, the Grinsztajn benchmark and TabArena~\citep{bischl2021openml,grinsztajn2022trees,erickson2025tabarena}.
The three benchmarks overlap, since \numDedupDupNamesClf{} classification names and \numDedupDupNamesReg{} regression names appear under two or three of them as distinct OpenML datasets, curations of one source (Table~\ref{tab:datasets}).
Grinsztajn itself lists \numGrinsztajnTwiceNames{} names under two of its suites; the evaluation keys its records by benchmark, name and task and, before any scoring, kept the later entry of each pair, the one with the higher OpenML id, so within a benchmark every count below has one entry per name.
Across benchmarks both entries of a shared name are scored, and the count with one entry per name keeps the earliest benchmark's (\S\ref{sec:final}); \numDedupBothRules{} names, diamonds and electricity, fall under both rules, the higher id within Grinsztajn and then the earliest benchmark across them.
Each dataset is scored in context with up to \numSuiteCtx{} context rows and \numSuiteKfeat{} features, by accuracy for classification and $R^2$ for regression, with no per-dataset training.
Each score is the mean over \numSuiteNSplits{} random halves of the dataset, with at most \numSuiteTestCap{} test rows scored per half and the \numSuiteKfeat{} features chosen by an F-test on the whole dataset before the split, the same for every model (\S\ref{app:protocol}).
On those scores two checkpoints are compared by per-dataset wins and losses, written wins:losses, with an exact two-sided sign test, overall, by class stratum and by benchmark.
Exact ties are dropped from the count and from the test, so the two numbers need not sum to the number of datasets, and the final classification pair has \numPairAllTies{} of them while regression has none.
The sign test weighs every dataset equally, so a large margin on one dataset counts no more than a small one, and the mean scores beside each count carry the magnitude the count discards (Table~\ref{tab:pair}).
Such a count has a floor, measured against a second run of \arm{ds} with the same recipe and the same seed, whose configuration differs only in the step cap and the stop rule, neither of which acts before the step compared, so the two diverge only through run-to-run non-determinism, not a new seed (\S\ref{app:config}).
Scored against its original at the same step it gives \numFloorClf{} on classification and \numFloorReg{} on regression, a floor that another seed could only widen; a win count inside it says nothing, and one outside it clears at least the noise of a re-run.
Gradient-boosted trees (HistGB in the tables and figures) and the linear model of each task, logistic regression on classification and ridge regression on regression, are fitted per dataset on its full training half, which on large datasets holds many times the \numSuiteCtx{} rows either arm reads, and serve as reference points.
Together these choices make every comparison below a paired one, in which each dataset is scored once per checkpoint under the same budget and a difference counts only when it clears at least the floor of a same-seed re-run (\S\ref{sec:final}).

\subsection{Convergence under the plateau rule}
\label{sec:convergence}

% ¶ res-watch

\begin{figure}[t]
\centering
\includegraphics[width=\linewidth]{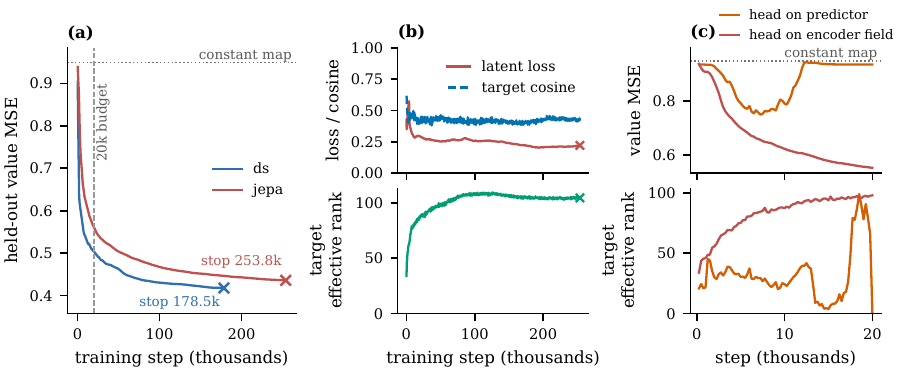}
\caption{\textbf{Held-out curves to the plateau.} (a) Value MSE of both arms on the held-out prior tables at every validation, up to each arm's plateau stop; the dotted line is the constant map and the dashed line the fixed budget of our earlier comparisons. (b) The \arm{jepa} arm's latent loss and target cosine (top) and target effective rank (bottom) on the same tables. (c) The fixed-budget pair behind the earlier collapse, the same recipe with the value head on the predictor or on the encoder field, one configuration key apart: value MSE (top) and target effective rank (bottom) on the same held-out tables as (b).}
\label{fig:curves}
\end{figure}

Each arm was stopped by one number, the squared error of the value head on the hidden cells of held-out prior tables (Figure~\ref{fig:curves}a).
That error is taken between the head's point prediction and the cell value clamped to the head's support, averaged over the hidden cells of tables drawn from a fixed seed disjoint from the training stream.
The held-out set holds \numDsValTables{} tables for \arm{ds} and \numJepaValTables{} for \arm{jepa}, the former being the first half of the latter, since both trainers draw the same stream from the same seed and each validates on its own default number of batches (\S\ref{app:config}).
The rule therefore never saw the real-data suite, and every suite score below is of a checkpoint chosen without it (\S\ref{sec:setup}).
For \arm{jepa} the rule read that same value MSE, so the latent loss and the collapse diagnostics in Figure~\ref{fig:curves}b were recorded at every validation but never stopped a run.
Converged therefore means converged in the value error, and the latent loss of \arm{jepa} had stopped falling by then, changing by \numJepaLatentLastWindow{} over the final patience window of \numPatienceSteps{} steps (Figure~\ref{fig:curves}b).
The latent loss could not have served as the stop signal, since a collapsed encoder drives it down as surely as a trained one, which is the failure \S\ref{sec:related} describes and Figure~\ref{fig:curves}c shows.
The value curve, by contrast, shows a collapse plainly: no constant output of the head can score below the variance of the held-out cells, \numJepaConstMap{} on the \arm{jepa} set, so the dotted line in Figure~\ref{fig:curves}a is what a collapsed encoder would score.

% ¶ res-convergence
The stop records and the trajectory together separate a slowdown from a ceiling, since \arm{jepa} reaches a plateau of its own, later, and its deficit against the final \arm{ds} checkpoint shrinks along the way (Figure~\ref{fig:curves}, Table~\ref{tab:trajectory}).
\arm{ds} stopped at \numDsStopStep{} steps after \numDsHours{} hours, with a final validation MSE of \numDsFinalValMse{} (Figure~\ref{fig:curves}a).
\arm{jepa} stopped at \numJepaStopStep{} steps after \numJepaHours{} hours, at \numJepaFinalValMse{}, which is \numJepaStepRatio{} times the steps and \numJepaHoursRatio{} times the wall-clock of \arm{ds}.
Both stop steps include the \numPatienceSteps{} steps of patience after each arm's last improvement by more than \numStopDelta{}, so the ratio of the steps at which \arm{jepa} and \arm{ds} last improved is \numJepaBestStepRatio{}.
Those stop points sit far past the step budget of our earlier comparisons, where the validation MSE of \arm{ds} stood at \numDsValMseTwentyk{} against \numDsFinalValMse{} at its plateau.
A fixed horizon had therefore scored an unconverged model, and any deficit measured there mixed a slowdown with whatever ceiling exists (the dashed line in Figure~\ref{fig:curves}a).
The latent term of \arm{jepa} survived to the plateau (Figure~\ref{fig:curves}b).
The validation latent loss of \arm{jepa} ended at \numJepaValJepa{}, with target cosine \numJepaTgtCos{} and target effective rank \numJepaErank{}.
The value curve of \arm{jepa} sat below the constant map from the first validation at step \numJepaFirstBelow{} and never returned above it (Figure~\ref{fig:curves}a).
Scored against \arm{ds} at the same step, \arm{jepa} trailed throughout, from \numTrajSameFirst{} at step \numTrajSameFirstStep{} to \numTrajSameLast{} at step \numTrajSameLastStep{} (Table~\ref{tab:trajectory} in the appendix).
Against the final \arm{ds} checkpoint instead, the deficit of \arm{jepa} narrowed from \numTrajFinalFirst{} at step \numTrajFinalFirstStep{} to \numTrajFinalLast{} at step \numTrajFinalLastStep{} without closing.
So the deficit at any fixed step reads mostly as lag, \arm{jepa} reaching at a later step what \arm{ds} had at an earlier one, and only what remains at both plateaus bounds the cost of the term from above (\S\ref{sec:final}).

\subsection{Converged comparison}
\label{sec:final}

% ¶ res-final

\input{generated/tab_bench.tex}

\begin{figure}[t]
\centering
\includegraphics[width=\linewidth]{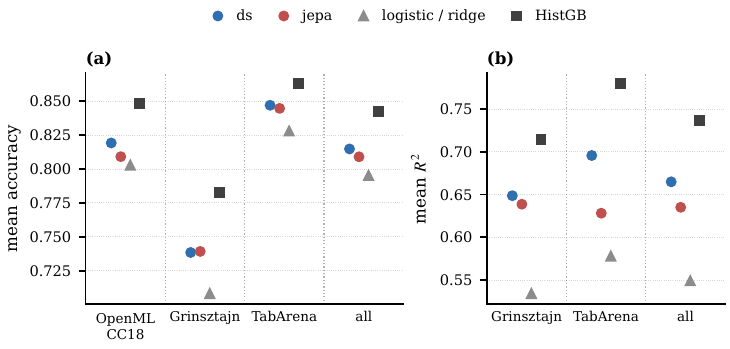}
\caption{\textbf{Benchmark means as points.} The entries of Table~\ref{tab:bench}, one marker per model: (a) accuracy on the classification sets of each benchmark and over all of them, (b) $R^2$ on the regression sets.}
\label{fig:bench}
\end{figure}

On all \numNClf{} classification datasets the final \arm{jepa} checkpoint wins \numPairAll{} against the final \arm{ds} checkpoint ($p$ \numPairAllP{}), with mean accuracy \numPairAllMeanJepa{} against \numPairAllMeanDs{} (Table~\ref{tab:pair} in the appendix).
On AUROC, stored beside accuracy for every classification dataset, the same pair gives \numPairAllAuc{} with \numPairAllAucTies{} ties, the same direction as the accuracy count.
With one entry kept per dataset name, the earliest benchmark's, the pool falls to \numDedupNClf{} classification and \numDedupNReg{} regression datasets and the counts become \numDedupPairAll{} (p \numDedupPairAllP{}) and \numDedupPairReg{} (p \numDedupPairRegP{}), the same direction at every stratum (Table~\ref{tab:pair}).
Keeping the latest benchmark's entry instead moves no count by more than \numDedupRuleMaxShift{} (\S\ref{sec:setup}).
The strata are ordered by what the head must resolve, from two classes through a few and many to a continuous target (Table~\ref{tab:pair}).
On binary tasks the count is \numPairBinary{} ($p$ \numPairBinaryP{}), significant, with a mean gap of \numPairBinaryGap{} accuracy.
On tasks with three to five classes it is \numPairFewclass{} ($p$ \numPairFewclassP{}), the one class stratum where the count is not significant.
The deficit is largest on the many-class stratum, six to ten classes, \numPairManyclass{} ($p$ \numPairManyclassP{}), and on regression, \numPairReg{} ($p$ \numPairRegP{}), with mean gaps of \numPairManyclassGap{} accuracy and \numPairRegGap{} $R^2$.
Across all strata, the mean gap at convergence is \numPairAllGap{} accuracy, so the deficit is small even where it is significant.
Table~\ref{tab:bench} and Figure~\ref{fig:bench} break the means down by benchmark, and on every block both arms score above the linear model and below the trees.
The largest gap between the arms is \numBenchArmGapMax{} $R^2$ on TabArena regression, against at most \numBenchArmGapMaxClf{} accuracy on any classification block (Table~\ref{tab:bench}).
That gap is concentrated rather than spread, since two datasets, \numRegTopNameOne{} and \numRegTopNameTwo{}, carry \numRegTopTwoShareTabarena{}\,\% of it with gaps of \numRegTopGapOne{} and \numRegTopGapTwo{} $R^2$ (Table~\ref{tab:datasets}).
The median gap over the \numNReg{} regression datasets is \numRegGapMedian{} $R^2$, and \arm{jepa} scores below ridge on \numRegJepaBelowRidge{} of them against \numRegDsBelowRidge{} for \arm{ds}.
By benchmark the classification count is significant only on OpenML-CC18, \numPairOpenmlClf{} ($p$ \numPairOpenmlClfP{}), while the regression counts on Grinsztajn, \numPairGrinsztajnReg{} ($p$ \numPairGrinsztajnRegP{}), and TabArena, \numPairTabarenaReg{} ($p$ \numPairTabarenaRegP{}), fall short of significance at their sizes (Table~\ref{tab:pair}).
Every block leans the same way, so the deficit is a property of the arm rather than of one benchmark's datasets, and only the largest block has the size to make it significant (Table~\ref{tab:pair}).
The stratum and benchmark counts are several tests on one pair of checkpoints, reported uncorrected as a breakdown of the overall count rather than as independent findings (Table~\ref{tab:pair}).
A stratum near the threshold is therefore read as a direction and not as a result (Table~\ref{tab:pair}).

\subsection{Suite score along training}
\label{sec:baselines}

% ¶ res-baselines

\begin{figure}[t]
\centering
\includegraphics[width=\linewidth]{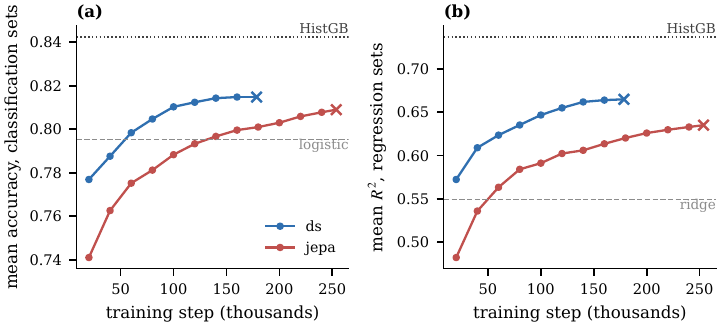}
\caption{\textbf{Real-data suite along training.} Mean score of every scored snapshot of both arms on the real-data suite, (a) accuracy over the classification sets and (b) $R^2$ over the regression sets, with the final checkpoints marked by a cross; the dotted line is gradient-boosted trees and the dashed line the linear model of each task, both fitted per dataset.}
\label{fig:suite}
\end{figure}

Neither arm beats gradient-boosted trees, with \arm{ds} at \numDsVsHistgbClf{} on classification and \numDsVsHistgbReg{} on regression and \arm{jepa} at \numJepaVsHistgbClf{} and \numJepaVsHistgbReg{} (Figure~\ref{fig:suite}).
Both arms beat the linear models, \arm{ds} at \numDsVsLogregClf{} and \numDsVsRidgeReg{} and \arm{jepa} at \numJepaVsLogregClf{} and \numJepaVsRidgeReg{}, the classification win rate of \arm{jepa} being the lowest of the four.
The distance to the trees, \numHistgbClfMean{} mean accuracy against \numDsClfMean{} for \arm{ds}, is several times the distance between the two arms.
That distance is not a like-for-like one, since the trees and the linear model read the full training half of each dataset while both arms read at most \numSuiteCtx{} of its rows, so it bounds nothing about what an objective could buy (\S\ref{sec:setup}).
Along training, the mean of \arm{ds} passes the logistic model by step \numDsCrossClf{} and that of \arm{jepa} by step \numJepaCrossClf{}, and on regression the two arms stand above ridge from steps \numDsCrossReg{} and \numJepaCrossReg{} on (Figure~\ref{fig:suite}).
So \arm{jepa} arrives later at each crossing while both arms end between the same two baselines: what differs is when the arm arrives, not where (Figure~\ref{fig:suite}).

%% file: generated/tab_bench.tex
% GENERATED by paper_v5/tools/build_tables.py -- do not edit by hand.
\begin{table}[t]
\centering
\small
\caption{\textbf{Final checkpoints by benchmark.} Mean score of both arms, of the linear model of each task and of gradient-boosted trees on the datasets of each benchmark and over all of them: accuracy for classification, $R^2$ for regression. Counts and sign tests are in Table~\ref{tab:pair}, per-dataset scores in Table~\ref{tab:datasets}.}
\label{tab:bench}
\begin{tabular}{lrrrrrrr}
\toprule
 & \multicolumn{4}{c}{classification (accuracy)} & \multicolumn{3}{c}{regression ($R^2$)} \\
\cmidrule(lr){2-5} \cmidrule(lr){6-8}
model & OpenML-CC18 & Grinsztajn & TabArena & all & Grinsztajn & TabArena & all \\
\midrule
\arm{ds} & .819 & .738 & .847 & .815 & .649 & .696 & .665 \\
\arm{jepa} & .809 & .739 & .845 & .809 & .639 & .628 & .635 \\
logistic / ridge & .803 & .708 & .828 & .795 & .534 & .578 & .549 \\
HistGB & .849 & .783 & .863 & .842 & .714 & .780 & .737 \\
\bottomrule
\end{tabular}
\end{table}

%% file: sections/discussion.tex
% GENERATED from paper_v5/{structure,claims,filler}.json by paper_v5/tools/build_prose.py -- do not edit.

\section{Discussion}
\label{sec:discussion}

% ¶ disc-findings
The recipe trains, since with the head on the encoder field the latent term neither collapsed nor drifted back toward the constant map over the whole run (\S\ref{sec:convergence}).
Yet it buys no accuracy on the real-data suite, since at convergence \arm{jepa} trails \arm{ds} on every stratum and benchmark where the count is significant and never leads the count where it is not (\S\ref{sec:final}).
Most of that deficit, as reported at a fixed budget, was a slowdown, and what remains at the plateau is small (\S\ref{sec:convergence}).
For the prescription this means that latent prediction beside the value loss on a tabular prior is no longer blocked by collapse, and that its case must now be made on something other than in-context accuracy (\S\ref{sec:future}).

% ¶ disc-why
Three features of the two arms can explain the deficit that remains.
The gradient of the latent term reaches the encoder through the predictor, so the encoder field that the value head reads is shaped by both losses (\S\ref{sec:method}).
Beyond that shared field, the two arms differ in masking policy, in the context split and in the warm-up, so the measured deficit bounds the cost of the latent term from above rather than estimating it, on the premise that none of the three helps the arm that carries the term, which this study does not test (\S\ref{sec:method}).
Where the deficit is largest, on many-class and regression tasks, the value head must resolve finer targets, which is consistent with a representation budget shared with the latent term (\S\ref{sec:final}).
Of the three, only the shared field belongs to the recipe itself, while the other two are choices of this study (\S\ref{sec:threats}).

\subsection{Threats to validity}
\label{sec:threats}

% ¶ disc-threats
One seed per arm is the sharpest limit on every claim, because the sign test spans datasets rather than seeds and the noise floor comes from a second run of the same seed, which a second seed would be expected to widen (\S\ref{sec:setup}).
What one seed does allow is the paired reading above, since both arms saw the same prior stream and the same suite, so the count compares two training runs and not yet two recipes (\S\ref{sec:setup}).
Next, no matched control with the latent weight set to zero was trained, so the deficit cannot be attributed to the latent term alone (\S\ref{sec:method}).
Three lesser limits remain.
One model size was trained, so the gap to gradient-boosted trees may be a size ceiling, a prior ceiling or the \numSuiteCtx{}-row context that caps what either arm reads of a dataset, three causes this study cannot separate (\S\ref{sec:baselines}).
The two arms share a learning rate, and \arm{jepa} adds a masking warm-up, an \ac{EMA} decay and one weight per loss, all carried over from earlier runs rather than tuned for this pair, so the bound of \S\ref{sec:final} holds under that setting rather than under each arm's best (\S\ref{sec:method}).
The suite likewise scores every dataset at one context and feature budget, so behaviour at other budgets is not measured (\S\ref{sec:setup}).

\subsection{Future work}
\label{sec:future}

% ¶ disc-future
A matched control with the latent weight at zero would turn the upper bound into an estimate (\S\ref{sec:threats}).
Without new training, whether the latent term buys robustness rather than accuracy, at short contexts or under noise, is testable on the same checkpoints with the same suite (\S\ref{sec:setup}).
That question is the one the prescription itself raises, since a latent target may discard what a value target must reproduce, and noise is where such a difference would show (\S\ref{sec:intro}).
A probe on the frozen \arm{jepa} field would test the prescription's two-stage form on these checkpoints as well (\S\ref{sec:method}).
Further out, the predictor allows filling hidden cells in latent space and chaining such fills, a use the value-only arm lacks and the present evaluation does not exercise (\S\ref{sec:method}).

%% file: sections/conclusion.tex
% GENERATED from paper_v5/{structure,claims,filler}.json by paper_v5/tools/build_prose.py -- do not edit.

\section{Conclusion}
\label{sec:conclusion}

% ¶ concl
We asked whether the latent-prediction prescription can be followed on a tabular foundation-model prior.
Its objective can be trained beside the value loss, once the head is moved off the predictor onto the encoder field, with the target and the mask tokens held fixed, and the term then reaches a plateau without collapse.
Yet it does not pay in real-data accuracy at convergence, trailing \arm{ds} in this run by a small margin on binary tasks and by more on many-class and regression tasks.
The lesson for comparisons of this kind is to stop each arm at its plateau rather than at a step count, since a fixed budget scores a slowdown as a ceiling.
The next step is the matched control that turns the measured bound into an estimate.

%% file: sections/appendix.tex
% GENERATED from paper_v5/{structure,claims,filler}.json by paper_v5/tools/build_prose.py -- do not edit.

\section{Additional material}
\label{app:material}

\subsection{Configurations}
\label{app:config}

% ¶ app-config
The two arms differ in the configuration keys of the two configurations that name the latent term and its target, the context split, the masking policy and the warm-up, and in the number of held-out batches each trainer validates on by default, \numDsValTables{} tables for \arm{ds} against \numJepaValTables{} for \arm{jepa} (\S\ref{sec:convergence}).
The plateau rule is one function shared by both trainers, so the two arms stopped under the same criterion.

\subsection{Collapse across priors}
\label{app:collapse}

\begin{figure}[htb]
\centering
\includegraphics[width=\linewidth]{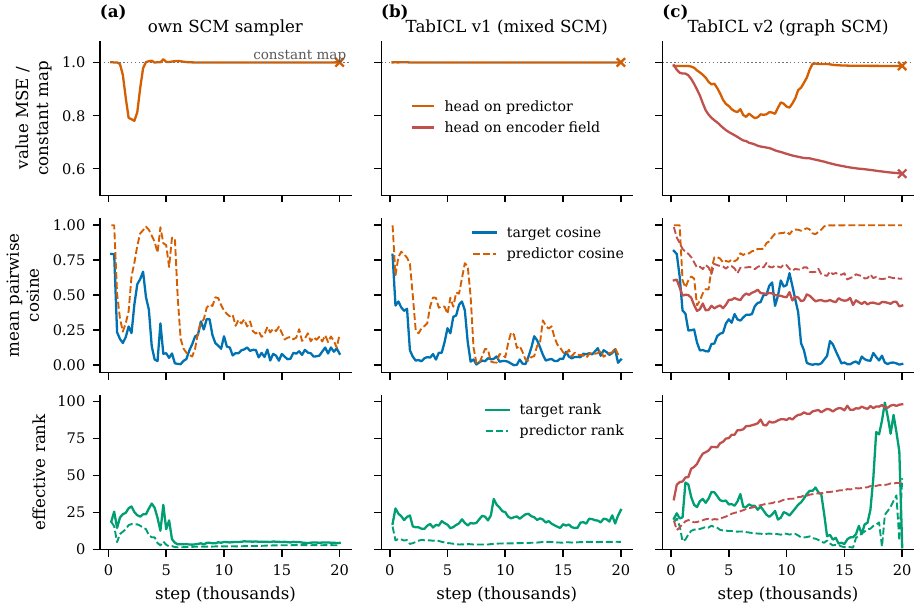}
\caption{\textbf{When and how the predictor-read wiring failed.} The runs of Table~\ref{tab:collapse}, one column per prior, at every validation: value MSE over the run's own constant map (top), the target field (solid) and the predictor output (dashed), for the mean pairwise cosine over the held-out cells (middle) and for the effective rank (bottom). In (c) the encoder-read run on the same prior is overlaid in the colour of Figure~\ref{fig:curves}c. Late in (c) the predictor-read run's target field is nearly constant and its rank is not stable, while the encoder-read overlay's rank keeps rising.}
\label{fig:collapse}
\end{figure}

% ¶ app-collapse
Table~\ref{tab:collapse} lists the three fixed-budget runs with the value head on the predictor, one per prior, beside the encoder-read run of Figure~\ref{fig:curves}c.
The three priors are our own \ac{SCM} sampler, the first TabICL generator, which mixes \ac{MLP} and tree \ac{SCM} engines, and the graph \ac{SCM} generator of TabICL v2 that the main runs use (\S\ref{sec:method}).
On every prior the predictor-read run ended within \numCpPredConstGapPctMax{}\,\% of its constant map with a target rank of at most \numCpPredErankMax{}, and scored \numCpPredClfLo{} to \numCpPredClfHi{} mean accuracy and at most \numCpPredRegMax{} mean $R^2$ on the suite, an $R^2$ no better than a constant prediction, against \numCpEncClf{} and \numCpEncReg{} for the encoder-read run at the same budget.
Those three configurations differ only in the prior, and the encoder-read run differs from the one on this prior in the single key that moves the head, so what the runs separate is the wiring, not the generator (\S\ref{app:config}).
Figure~\ref{fig:collapse} draws the same runs at every validation: the value error over the constant map, the mean pairwise cosine of the predictor output and of the target field, and the effective rank of the target field and of the predictor output.
On the two priors where the value head learned at all, the predictor output flattened first, its cosine crossing \numCpRulePredCosFlat{} at steps \numCpScmPredFlatStep{} and \numCpTabiclTwoPredFlatStep{}, after the value error's minimum at \numCpScmMinStep{} and \numCpTabiclTwoMinStep{} and before its return to \numCpRuleBackFrac{} of the constant map at \numCpScmBackStep{} and \numCpTabiclTwoBackStep{}.
The target field lost rank only afterwards, its effective rank falling below \numCpRuleRank{} at steps \numCpScmRankStep{} and \numCpTabiclTwoRankStep{}.
On TabICL v1 the value error never left the constant map, its minimum being \numCpTabiclOneMinOverConst{} of the map, while the predictor output kept an effective rank of at most \numCpTabiclOnePredErankMax{} throughout.
So the head lost the value task while reading an output that had already gone flat, in the order \S\ref{sec:intro} gives, and the target field followed rather than led.

\input{generated/tab_collapse.tex}

\subsection{Evaluation protocol}
\label{app:protocol}

% ¶ app-protocol
The suite is scored by the evaluation scripts, which draw the three OpenML suites, give each model at most \numSuiteCtx{} rows of context and \numSuiteKfeat{} features picked by an F-test ahead of the split, average \numSuiteNSplits{} random halves scoring no more than \numSuiteTestCap{} test rows in each, and apply an exact two-sided sign test with ties dropped, as \S\ref{sec:setup} states.

\subsection{Snapshots along the trajectory}
\label{app:trajectory}

% ¶ app-trajectory
Table~\ref{tab:trajectory} lists, for every scored snapshot of \arm{jepa}, the two counts of \S\ref{sec:convergence}, one against the \arm{ds} snapshot of the same step and one against the final \arm{ds} checkpoint, of which the section quotes the first and the last.

\input{generated/tab_trajectory.tex}

\subsection{Sign tests by stratum and benchmark}
\label{app:strata}

% ¶ app-strata
Every count behind \S\ref{sec:final} is listed in Table~\ref{tab:pair} with its $p$ value, first by class stratum and then by benchmark.

\input{generated/tab_pair.tex}

\subsection{Per-dataset scores}
\label{app:datasets}

% ¶ app-datasets
Every number behind Table~\ref{tab:pair} and every entry of Table~\ref{tab:bench} except the tree means is listed in Table~\ref{tab:datasets}, one row per dataset, grouped by benchmark, with the linear model of each task beside the two arms.

\input{generated/tab_datasets.tex}

%% file: generated/tab_collapse.tex
% GENERATED by paper_v5/tools/build_tables.py -- do not edit by hand.
\begin{table}[htb]
\centering
\small
\caption{\textbf{The predictor-read wiring across priors.} Fixed-budget runs of \numPairBudget{} steps, one per prior, with the value head on the predictor, and for reference the run on the paper's prior whose head reads the encoder field (Figure~\ref{fig:curves}c). Value MSE is the held-out value error at the end of the run; the constant map is the variance of that run's own held-out target cells, the score of a collapsed encoder; the minimum is the lowest held-out value error along the run and its step; rank is the target effective rank at the end (ceiling \numEmb{}); suite means are accuracy over the \numNClf{} classification and $R^2$ over the \numNReg{} regression datasets.}
\label{tab:collapse}
\setlength{\tabcolsep}{4pt}
\begin{tabular}{llrrrrrr}
\toprule
prior & head reads & value MSE & constant map & minimum (step) & rank & suite acc. & suite $R^2$ \\
\midrule
own SCM sampler & predictor & .971 & .971 & .757 (2,250) & 4.4 & .585 & -.003 \\
TabICL v1 (mixed SCM) & predictor & 1.014 & 1.014 & 1.014 (1,750) & 26.7 & .582 & -.003 \\
TabICL v2 (graph SCM) & predictor & .936 & .949 & .750 (7,250) & 1.5 & .570 & -.003 \\
\midrule
TabICL v2 (graph SCM) & encoder & .551 & .949 & .551 (20,000) & 98.1 & .750 & .490 \\
\bottomrule
\end{tabular}
\end{table}

%% file: generated/tab_trajectory.tex
% GENERATED by paper_v5/tools/build_tables.py -- do not edit by hand.
\begin{table}[t]
\centering
\small
\caption{\textbf{Snapshots of \arm{jepa} along its trajectory.} Per-dataset wins against \arm{ds} at the same step and against its final checkpoint, as wins and losses for all classification sets, the many-class stratum and regression; a dash marks a step past the stop of \arm{ds}.}
\label{tab:trajectory}
\begin{tabular}{rrrrrrr}
\toprule
step & \multicolumn{3}{c}{vs.\ \arm{ds} at the same step} & \multicolumn{3}{c}{vs.\ \arm{ds} final} \\
 & all & 6--10 & reg & all & 6--10 & reg \\
\midrule
20,000 & 22:80 & 1:18 & 7:25 & 9:96 & 0:19 & 0:32 \\
40,000 & 23:81 & 3:16 & 3:29 & 6:96 & 1:17 & 0:32 \\
60,000 & 26:79 & 3:16 & 2:30 & 15:87 & 2:17 & 0:32 \\
80,000 & 20:86 & 1:18 & 3:28 & 11:91 & 1:18 & 1:31 \\
100,000 & 16:86 & 2:17 & 4:28 & 17:89 & 1:18 & 3:29 \\
120,000 & 15:89 & 1:17 & 2:30 & 18:87 & 1:18 & 2:30 \\
140,000 & 19:83 & 2:17 & 4:28 & 19:81 & 1:17 & 3:29 \\
160,000 & 17:79 & 1:18 & 4:28 & 22:81 & 1:18 & 4:28 \\
180,000 & -- & -- & -- & 21:82 & 1:18 & 2:29 \\
200,000 & -- & -- & -- & 24:78 & 2:16 & 1:30 \\
220,000 & -- & -- & -- & 36:68 & 2:16 & 5:27 \\
240,000 & -- & -- & -- & 31:69 & 2:16 & 6:26 \\
\bottomrule
\end{tabular}
\end{table}

%% file: generated/tab_pair.tex
% GENERATED by paper_v5/tools/build_tables.py -- do not edit by hand.
\begin{table}[htb]
\centering
\small
\caption{\textbf{Sign tests by stratum and benchmark.} Per-dataset wins and losses of \arm{jepa} against \arm{ds} at their final checkpoints, with the exact two-sided sign-test $p$, by class stratum and by benchmark. Means are accuracy (classification) and $R^2$ (regression). The last block repeats the class strata and the regression count with one entry kept per dataset name.}
\label{tab:pair}
\begin{tabular}{lrrrrrr}
\toprule
stratum & $n$ & \arm{jepa} W:L & ties & $p$ & \arm{jepa} mean & \arm{ds} mean \\
\midrule
all classification & 115 & 32:70 & 13 & .00021 & .809 & .815 \\
\quad binary & 78 & 24:44 & 10 & .02053 & .823 & .826 \\
\quad 3--5 classes & 18 & 6:11 & 1 & .33231 & .795 & .802 \\
\quad 6--10 classes & 19 & 2:15 & 2 & .00235 & .765 & .782 \\
\quad OpenML-CC18 & 59 & 13:42 & 4 & .00011 & .809 & .819 \\
\quad Grinsztajn & 19 & 7:11 & 1 & .48068 & .739 & .738 \\
\quad TabArena & 37 & 12:17 & 8 & .45826 & .845 & .847 \\
\midrule
all regression & 32 & 8:24 & 0 & .007 & .635 & .665 \\
\quad Grinsztajn & 21 & 6:15 & 0 & .07835 & .639 & .649 \\
\quad TabArena & 11 & 2:9 & 0 & .06543 & .628 & .696 \\
\midrule
\multicolumn{7}{l}{\emph{one entry per dataset name, the earliest benchmark's (\S\ref{sec:setup})}} \\
all classification & 102 & 29:63 & 10 & .00051 & .812 & .818 \\
\quad binary & 67 & 21:38 & 8 & .03634 & .827 & .830 \\
\quad 3--5 classes & 17 & 6:10 & 1 & .4545 & .792 & .797 \\
\quad 6--10 classes & 18 & 2:15 & 1 & .00235 & .777 & .796 \\
all regression & 30 & 8:22 & 0 & .01612 & .626 & .654 \\
\bottomrule
\end{tabular}
\end{table}

%% file: generated/tab_datasets.tex
% GENERATED by paper_v5/tools/build_tables.py -- do not edit by hand.
{\footnotesize
\begin{longtable}{lrrrrr}
\caption{\textbf{Per-dataset scores by benchmark.} Final checkpoints of both arms and the linear baseline of each task on every dataset of the suite, grouped by benchmark, each with its OpenML dataset id: accuracy with the number of classes $k$ for classification, $R^2$ for regression. The better of the two arms is in bold; each block ends with its mean and the count of datasets won by each arm.}
\label{tab:datasets} \\
\toprule
dataset & id & $k$ & \arm{ds} & \arm{jepa} & logistic / ridge \\
\midrule
\endfirsthead
\caption[]{\textbf{Per-dataset scores by benchmark} (continued).} \\
\toprule
dataset & id & $k$ & \arm{ds} & \arm{jepa} & logistic / ridge \\
\midrule
\endhead
\bottomrule
\endlastfoot
\multicolumn{6}{l}{\emph{OpenML-CC18, classification (59 datasets)}} \\
adult & 1590 & 2 & \textbf{.830} & .816 & .789 \\
analcatdata\_authorship & 458 & 4 & .990 & .990 & .988 \\
analcatdata\_dmft & 469 & 6 & .196 & \textbf{.211} & .204 \\
balance-scale & 11 & 3 & .837 & \textbf{.842} & .873 \\
bank-marketing & 1461 & 2 & \textbf{.897} & .891 & .902 \\
banknote-authentication & 1462 & 2 & \textbf{.998} & .997 & .992 \\
blood-transfusion-service-center & 1464 & 2 & .772 & .772 & .774 \\
breast-w & 15 & 2 & \textbf{.972} & .971 & .966 \\
car & 40975 & 4 & \textbf{.841} & .835 & .821 \\
churn & 40701 & 2 & .883 & \textbf{.889} & .856 \\
climate-model-simulation-crashes & 40994 & 2 & .915 & .915 & .923 \\
cmc & 23 & 3 & .522 & \textbf{.524} & .499 \\
connect-4 & 40668 & 3 & \textbf{.659} & .658 & .686 \\
credit-approval & 29 & 2 & .873 & \textbf{.883} & .867 \\
credit-g & 31 & 2 & \textbf{.726} & .713 & .738 \\
cylinder-bands & 6332 & 2 & .722 & \textbf{.728} & .696 \\
diabetes & 37 & 2 & \textbf{.779} & .768 & .773 \\
dna & 40670 & 3 & .926 & \textbf{.935} & .960 \\
dresses-sales & 23381 & 2 & \textbf{.601} & .591 & .555 \\
electricity & 151 & 2 & \textbf{.786} & .785 & .769 \\
eucalyptus & 188 & 5 & .601 & \textbf{.609} & .555 \\
first-order-theorem-proving & 1475 & 6 & \textbf{.503} & .487 & .484 \\
GesturePhaseSegmentationProcessed & 4538 & 5 & \textbf{.531} & .518 & .464 \\
ilpd & 1480 & 2 & \textbf{.702} & .688 & .714 \\
jm1 & 1053 & 2 & .805 & .805 & .799 \\
jungle\_chess\_2pcs\_raw\_endgame\_complete & 41027 & 3 & \textbf{.725} & .723 & .682 \\
kc1 & 1067 & 2 & .855 & \textbf{.860} & .856 \\
kc2 & 1063 & 2 & \textbf{.839} & .838 & .831 \\
kr-vs-kp & 3 & 2 & \textbf{.941} & .820 & .953 \\
mfeat-factors & 12 & 10 & \textbf{.935} & .912 & .953 \\
mfeat-fourier & 14 & 10 & \textbf{.824} & .795 & .801 \\
mfeat-karhunen & 16 & 10 & \textbf{.935} & .911 & .952 \\
mfeat-morphological & 18 & 10 & \textbf{.750} & .731 & .676 \\
mfeat-pixel & 40979 & 10 & \textbf{.870} & .813 & .909 \\
mfeat-zernike & 22 & 10 & \textbf{.777} & .722 & .792 \\
MiceProtein & 40966 & 8 & \textbf{.946} & .935 & .831 \\
nomao & 1486 & 2 & .917 & \textbf{.924} & .934 \\
numerai28.6 & 23517 & 2 & \textbf{.503} & .500 & .523 \\
optdigits & 28 & 10 & \textbf{.866} & .818 & .971 \\
ozone-level-8hr & 1487 & 2 & \textbf{.930} & .926 & .929 \\
pc1 & 1068 & 2 & .930 & \textbf{.932} & .917 \\
pc3 & 1050 & 2 & \textbf{.889} & .887 & .887 \\
pc4 & 1049 & 2 & \textbf{.892} & .889 & .902 \\
pendigits & 32 & 10 & \textbf{.951} & .934 & .951 \\
PhishingWebsites & 4534 & 2 & .936 & \textbf{.938} & .894 \\
phoneme & 1489 & 2 & .842 & \textbf{.844} & .751 \\
qsar-biodeg & 1494 & 2 & \textbf{.855} & .850 & .858 \\
satimage & 182 & 6 & \textbf{.883} & .876 & .845 \\
segment & 40984 & 7 & \textbf{.924} & .918 & .869 \\
semeion & 1501 & 10 & \textbf{.796} & .788 & .791 \\
sick & 38 & 2 & \textbf{.973} & .972 & .967 \\
spambase & 44 & 2 & \textbf{.938} & .926 & .921 \\
splice & 46 & 3 & \textbf{.926} & .904 & .916 \\
steel-plates-fault & 40982 & 7 & \textbf{.763} & .754 & .521 \\
tic-tac-toe & 50 & 2 & \textbf{.745} & .734 & .704 \\
vehicle & 54 & 4 & \textbf{.748} & .733 & .799 \\
wall-robot-navigation & 1497 & 4 & \textbf{.922} & .894 & .702 \\
wdbc & 1510 & 2 & \textbf{.959} & .943 & .957 \\
wilt & 40983 & 2 & \textbf{.979} & .975 & .969 \\
\emph{mean} & & & .819 & .809 & .803 \\
\emph{datasets won} & & & 42 & 13 & \\
\midrule
\multicolumn{6}{l}{\emph{Grinsztajn, classification (19 datasets)}} \\
albert & 45035 & 2 & \textbf{.626} & .614 & .631 \\
analcatdata\_supreme & 44055 & 10 & .975 & .975 & .676 \\
bank-marketing & 44126 & 2 & .758 & \textbf{.768} & .743 \\
Bioresponse & 45019 & 2 & \textbf{.702} & .699 & .693 \\
california & 45028 & 2 & \textbf{.847} & .846 & .838 \\
compas-two-years & 45039 & 2 & .648 & \textbf{.679} & .684 \\
credit & 44089 & 2 & .747 & \textbf{.760} & .717 \\
default-of-credit-card-clients & 45036 & 2 & \textbf{.727} & .721 & .693 \\
Diabetes130US & 45022 & 2 & .555 & \textbf{.573} & .588 \\
electricity & 44156 & 2 & \textbf{.774} & .766 & .740 \\
eye\_movements & 44157 & 2 & .519 & \textbf{.544} & .551 \\
heloc & 45026 & 2 & \textbf{.697} & .693 & .697 \\
house\_16H & 44123 & 2 & \textbf{.859} & .851 & .791 \\
jannis & 45021 & 2 & \textbf{.713} & .690 & .744 \\
MagicTelescope & 44125 & 2 & .841 & \textbf{.842} & .770 \\
MiniBooNE & 44128 & 2 & \textbf{.896} & .893 & .896 \\
pol & 44122 & 2 & \textbf{.932} & .927 & .872 \\
road-safety & 45038 & 2 & .670 & \textbf{.687} & .597 \\
wine\_quality & 44136 & 7 & \textbf{.544} & .518 & .534 \\
\emph{mean} & & & .738 & .739 & .708 \\
\emph{datasets won} & & & 11 & 7 & \\
\midrule
\multicolumn{6}{l}{\emph{Grinsztajn, regression (21 datasets)}} \\
abalone & 45042 & -- & \textbf{.527} & .516 & .490 \\
Ailerons & 44137 & -- & .657 & \textbf{.671} & .809 \\
Allstate\_Claims\_Severity & 45046 & -- & .388 & \textbf{.399} & .419 \\
Bike\_Sharing\_Demand & 44142 & -- & .627 & \textbf{.630} & .341 \\
Brazilian\_houses & 44141 & -- & \textbf{.997} & .995 & .821 \\
cpu\_act & 44132 & -- & \textbf{.873} & .863 & .767 \\
diamonds & 44140 & -- & \textbf{.942} & .940 & .933 \\
elevators & 44134 & -- & \textbf{.499} & .473 & .238 \\
house\_16H & 44139 & -- & \textbf{.641} & .624 & .348 \\
house\_sales & 44144 & -- & \textbf{.851} & .812 & .764 \\
houses & 44138 & -- & \textbf{.768} & .683 & .639 \\
medical\_charges & 45048 & -- & .977 & \textbf{.977} & .826 \\
Mercedes\_Benz\_Greener\_Manufacturing & 44061 & -- & \textbf{.497} & .490 & .483 \\
MiamiHousing2016 & 44147 & -- & \textbf{.891} & .873 & .728 \\
pol & 44133 & -- & .894 & \textbf{.901} & .469 \\
seattlecrime6 & 45043 & -- & \textbf{.163} & .157 & .129 \\
sulfur & 44145 & -- & \textbf{.536} & .533 & .368 \\
superconduct & 44148 & -- & \textbf{.835} & .828 & .724 \\
topo\_2\_1 & 45041 & -- & \textbf{.023} & .005 & .037 \\
visualizing\_soil & 44056 & -- & \textbf{.997} & .997 & .837 \\
yprop\_4\_1 & 45032 & -- & .038 & \textbf{.041} & .046 \\
\emph{mean} & & & .649 & .639 & .534 \\
\emph{datasets won} & & & 15 & 6 & \\
\midrule
\multicolumn{6}{l}{\emph{TabArena, classification (37 datasets)}} \\
Amazon\_employee\_access & 46905 & 2 & .948 & .948 & .948 \\
anneal & 46906 & 5 & \textbf{.978} & .973 & .939 \\
APSFailure & 46908 & 2 & .982 & .982 & .980 \\
bank-marketing & 46910 & 2 & .883 & \textbf{.884} & .891 \\
Bank\_Customer\_Churn & 46911 & 2 & .835 & \textbf{.840} & .801 \\
blood-transfusion-service-center & 46913 & 2 & .764 & \textbf{.774} & .774 \\
churn & 46915 & 2 & \textbf{.889} & .887 & .878 \\
coil2000\_insurance\_policies & 46916 & 2 & .945 & .945 & .944 \\
credit-g & 46918 & 2 & \textbf{.719} & .710 & .725 \\
credit\_card\_clients\_default & 46919 & 2 & .803 & \textbf{.812} & .811 \\
customer\_satisfaction\_in\_airline & 46920 & 2 & \textbf{.883} & .848 & .759 \\
diabetes & 46921 & 2 & \textbf{.760} & .759 & .766 \\
Diabetes130US & 46922 & 2 & .903 & .903 & .902 \\
E-CommereShippingData & 46924 & 2 & .648 & \textbf{.659} & .649 \\
Fitness\_Club & 46927 & 2 & \textbf{.776} & .767 & .771 \\
GiveMeSomeCredit & 46929 & 2 & \textbf{.935} & .934 & .936 \\
hazelnut-spread-contaminant-detection & 46930 & 2 & \textbf{.881} & .857 & .704 \\
heloc & 46932 & 2 & .701 & .701 & .711 \\
HR\_Analytics\_Job\_Change\_of\_Data\_Scientists & 46935 & 2 & \textbf{.768} & .757 & .756 \\
in\_vehicle\_coupon\_recommendation & 46937 & 2 & .626 & \textbf{.646} & .644 \\
Is-this-a-good-customer & 46938 & 2 & \textbf{.893} & .892 & .891 \\
jm1 & 46979 & 2 & \textbf{.805} & .805 & .796 \\
kddcup09\_appetency & 46939 & 2 & .982 & .982 & .982 \\
Marketing\_Campaign & 46940 & 2 & \textbf{.872} & .869 & .864 \\
maternal\_health\_risk & 46941 & 3 & \textbf{.776} & .755 & .613 \\
MIC & 46980 & 8 & .887 & \textbf{.893} & .868 \\
NATICUSdroid & 46969 & 2 & .915 & \textbf{.917} & .923 \\
online\_shoppers\_intention & 46947 & 2 & .901 & \textbf{.908} & .892 \\
polish\_companies\_bankruptcy & 46950 & 2 & .935 & \textbf{.938} & .937 \\
qsar-biodeg & 46952 & 2 & \textbf{.850} & .846 & .860 \\
SDSS17 & 46955 & 3 & \textbf{.967} & .961 & .872 \\
seismic-bumps & 46956 & 2 & .935 & .935 & .935 \\
splice & 46958 & 3 & \textbf{.882} & .840 & .844 \\
students\_dropout\_and\_academic\_success & 46960 & 3 & .737 & \textbf{.749} & .768 \\
taiwanese\_bankruptcy\_prediction & 46962 & 2 & \textbf{.969} & .968 & .963 \\
website\_phishing & 46963 & 3 & .869 & \textbf{.871} & .816 \\
wine\_quality & 46964 & 7 & .539 & .539 & .519 \\
\emph{mean} & & & .847 & .845 & .828 \\
\emph{datasets won} & & & 17 & 12 & \\
\midrule
\multicolumn{6}{l}{\emph{TabArena, regression (11 datasets)}} \\
airfoil\_self\_noise & 46904 & -- & \textbf{.789} & .633 & .282 \\
Another-Dataset-on-used-Fiat-500 & 46907 & -- & .840 & \textbf{.841} & .842 \\
concrete\_compressive\_strength & 46917 & -- & \textbf{.792} & .446 & .571 \\
diamonds & 46923 & -- & \textbf{.883} & .862 & .854 \\
Food\_Delivery\_Time & 46928 & -- & \textbf{.238} & .225 & .202 \\
healthcare\_insurance\_expenses & 46931 & -- & \textbf{.765} & .718 & .741 \\
houses & 46934 & -- & \textbf{.774} & .680 & .638 \\
miami\_housing & 46942 & -- & \textbf{.796} & .778 & .690 \\
physiochemical\_protein & 46949 & -- & \textbf{.388} & .356 & .278 \\
QSAR\_fish\_toxicity & 46954 & -- & \textbf{.584} & .564 & .555 \\
superconductivity & 46961 & -- & .804 & \textbf{.807} & .701 \\
\emph{mean} & & & .696 & .628 & .578 \\
\emph{datasets won} & & & 9 & 2 & \\
\end{longtable}}

%% file: main.bbl
\begin{thebibliography}{34}
\providecommand{\natexlab}[1]{#1}
\providecommand{\url}[1]{\texttt{#1}}
\expandafter\ifx\csname urlstyle\endcsname\relax
  \providecommand{\doi}[1]{doi: #1}\else
  \providecommand{\doi}{doi: \begingroup \urlstyle{rm}\Url}\fi

\bibitem[Assran et~al.(2023)Assran, Duval, Misra, Bojanowski, Vincent, Rabbat,
  LeCun, and Ballas]{assran2023ijepa}
Mahmoud Assran, Quentin Duval, Ishan Misra, Piotr Bojanowski, Pascal Vincent,
  Michael Rabbat, Yann LeCun, and Nicolas Ballas.
\newblock Self-supervised learning from images with a joint-embedding
  predictive architecture.
\newblock In \emph{Proceedings of the IEEE/CVF Conference on Computer Vision
  and Pattern Recognition (CVPR)}, 2023.

\bibitem[Baevski et~al.(2022{\natexlab{a}})Baevski, Babu, Hsu, and
  Auli]{baevski2022data2vec2}
Alexei Baevski, Arun Babu, Wei-Ning Hsu, and Michael Auli.
\newblock Efficient self-supervised learning with contextualized target
  representations for vision, speech and language.
\newblock \emph{arXiv preprint arXiv:2212.07525}, 2022{\natexlab{a}}.

\bibitem[Baevski et~al.(2022{\natexlab{b}})Baevski, Hsu, Xu, Babu, Gu, and
  Auli]{baevski2022data2vec}
Alexei Baevski, Wei-Ning Hsu, Qiantong Xu, Arun Babu, Jiatao Gu, and Michael
  Auli.
\newblock {data2vec}: A general framework for self-supervised learning in
  speech, vision and language.
\newblock In \emph{Proceedings of the 39th International Conference on Machine
  Learning (ICML)}. PMLR, 2022{\natexlab{b}}.

\bibitem[Bahri et~al.(2022)Bahri, Jiang, Tay, and Metzler]{bahri2022scarf}
Dara Bahri, Heinrich Jiang, Yi~Tay, and Donald Metzler.
\newblock {SCARF}: Self-supervised contrastive learning using random feature
  corruption.
\newblock In \emph{International Conference on Learning Representations
  (ICLR)}, 2022.
\newblock Spotlight.

\bibitem[Balestriero \& LeCun(2025)Balestriero and
  LeCun]{balestriero2025lejepa}
Randall Balestriero and Yann LeCun.
\newblock {LeJEPA}: Provable and scalable self-supervised learning without the
  heuristics.
\newblock \emph{arXiv preprint arXiv:2511.08544}, 2025.

\bibitem[Bardes et~al.(2022)Bardes, Ponce, and LeCun]{bardes2022vicreg}
Adrien Bardes, Jean Ponce, and Yann LeCun.
\newblock {VICReg}: Variance-invariance-covariance regularization for
  self-supervised learning.
\newblock In \emph{International Conference on Learning Representations
  (ICLR)}, 2022.

\bibitem[Bardes et~al.(2024)Bardes, Garrido, Ponce, Chen, Rabbat, LeCun,
  Assran, and Ballas]{bardes2024vjepa}
Adrien Bardes, Quentin Garrido, Jean Ponce, Xinlei Chen, Michael Rabbat, Yann
  LeCun, Mahmoud Assran, and Nicolas Ballas.
\newblock Revisiting feature prediction for learning visual representations
  from video.
\newblock \emph{arXiv preprint arXiv:2404.08471}, 2024.

\bibitem[Bischl et~al.(2021)Bischl, Casalicchio, Feurer, Hutter, Lang,
  Mantovani, van Rijn, and Vanschoren]{bischl2021openml}
Bernd Bischl, Giuseppe Casalicchio, Matthias Feurer, Frank Hutter, Michel Lang,
  Rafael~G. Mantovani, Jan~N. van Rijn, and Joaquin Vanschoren.
\newblock {OpenML} benchmarking suites.
\newblock In \emph{Advances in Neural Information Processing Systems (NeurIPS)
  Datasets and Benchmarks Track}, 2021.

\bibitem[Bouadi et~al.(2026)Bouadi, Bouarour, Kulkarni, Dubey, Tanna, and
  Sankarapu]{bouadi2026shapingprior}
Mohamed Bouadi, Nassim Bouarour, Varun Kulkarni, Shivam Dubey, Aditya Tanna,
  and Vinay~Kumar Sankarapu.
\newblock Shaping the prior: How synthetic task distributions determine tabular
  foundation model quality.
\newblock \emph{arXiv preprint arXiv:2605.18971}, 2026.

\bibitem[Chen \& He(2020)Chen and He]{chen2020simsiam}
Xinlei Chen and Kaiming He.
\newblock Exploring simple siamese representation learning, 2020.

\bibitem[Erickson et~al.(2025)Erickson, Purucker, Tschalzev, Holzm\"{u}ller,
  Desai, Salinas, and Hutter]{erickson2025tabarena}
Nick Erickson, Lennart Purucker, Andrej Tschalzev, David Holzm\"{u}ller,
  Prateek~Mutalik Desai, David Salinas, and Frank Hutter.
\newblock {TabArena}: A living benchmark for machine learning on tabular data.
\newblock \emph{arXiv preprint arXiv:2506.16791}, 2025.

\bibitem[Garrido et~al.(2023)Garrido, Balestriero, Najman, and
  LeCun]{garrido2023rankme}
Quentin Garrido, Randall Balestriero, Laurent Najman, and Yann LeCun.
\newblock {RankMe}: Assessing the downstream performance of pretrained
  self-supervised representations by their rank.
\newblock In \emph{Proceedings of the 40th International Conference on Machine
  Learning (ICML)}. PMLR, 2023.

\bibitem[Grill et~al.(2020)Grill, Strub, Altch\'{e}, Tallec, Richemond,
  Buchatskaya, Doersch, Pires, Guo, Azar, Piot, Kavukcuoglu, Munos, and
  Valko]{grill2020byol}
Jean-Bastien Grill, Florian Strub, Florent Altch\'{e}, Corentin Tallec,
  Pierre~H. Richemond, Elena Buchatskaya, Carl Doersch, Bernardo~Avila Pires,
  Zhaohan~Daniel Guo, Mohammad~Gheshlaghi Azar, Bilal Piot, Koray Kavukcuoglu,
  R\'{e}mi Munos, and Michal Valko.
\newblock Bootstrap your own latent: A new approach to self-supervised
  learning.
\newblock In \emph{Advances in Neural Information Processing Systems 33
  (NeurIPS)}, 2020.

\bibitem[Grinsztajn et~al.(2022)Grinsztajn, Oyallon, and
  Varoquaux]{grinsztajn2022trees}
L\'{e}o Grinsztajn, Edouard Oyallon, and Ga\"{e}l Varoquaux.
\newblock Why do tree-based models still outperform deep learning on typical
  tabular data?
\newblock In \emph{Advances in Neural Information Processing Systems (NeurIPS)
  Datasets and Benchmarks Track}, 2022.

\bibitem[Hollmann et~al.(2023)Hollmann, M\"{u}ller, Eggensperger, and
  Hutter]{hollmann2023tabpfn}
Noah Hollmann, Samuel M\"{u}ller, Katharina Eggensperger, and Frank Hutter.
\newblock {TabPFN}: A transformer that solves small tabular classification
  problems in a second.
\newblock In \emph{International Conference on Learning Representations
  (ICLR)}, 2023.

\bibitem[Hollmann et~al.(2025)Hollmann, M\"{u}ller, Purucker, Krishnakumar,
  K\"{o}rfer, Hoo, Schirrmeister, and Hutter]{hollmann2025tabpfnv2}
Noah Hollmann, Samuel M\"{u}ller, Lennart Purucker, Arjun Krishnakumar, Max
  K\"{o}rfer, Shi~Bin Hoo, Robin~Tibor Schirrmeister, and Frank Hutter.
\newblock Accurate predictions on small data with a tabular foundation model.
\newblock \emph{Nature}, 637\penalty0 (8045):\penalty0 319--326, 2025.
\newblock \doi{10.1038/s41586-024-08328-6}.

\bibitem[Jing et~al.(2021)Jing, Vincent, LeCun, and Tian]{jing2021dimensional}
Li~Jing, Pascal Vincent, Yann LeCun, and Yuandong Tian.
\newblock Understanding dimensional collapse in contrastive self-supervised
  learning.
\newblock \emph{arXiv preprint arXiv:2110.09348}, 2021.

\bibitem[Kim et~al.(2024)Kim, Grinsztajn, and Varoquaux]{kim2024carte}
Myung~Jun Kim, L\'{e}o Grinsztajn, and Ga\"{e}l Varoquaux.
\newblock {CARTE}: Pretraining and transfer for tabular learning.
\newblock \emph{arXiv preprint arXiv:2402.16785}, 2024.

\bibitem[LeCun(2022)]{lecun2022path}
Yann LeCun.
\newblock A path towards autonomous machine intelligence, 2022.
\newblock Version 0.9.2, 2022-06-27. OpenReview position paper,
  \url{https://openreview.net/forum?id=BZ5a1r-kVsf}.

\bibitem[Littwin et~al.(2024)Littwin, Saremi, Advani, Thilak, Nakkiran, Huang,
  and Susskind]{littwin2024jepabias}
Etai Littwin, Omid Saremi, Madhu Advani, Vimal Thilak, Preetum Nakkiran, Chen
  Huang, and Joshua Susskind.
\newblock How {JEPA} avoids noisy features: The implicit bias of deep linear
  self distillation networks, 2024.

\bibitem[Ma et~al.(2025)Ma, Thomas, Hosseinzadeh, Kamkari, Labach, Cresswell,
  Golestan, Yu, Caterini, and Volkovs]{ma2025tabdpt}
Junwei Ma, Valentin Thomas, Rasa Hosseinzadeh, Hamidreza Kamkari, Alex Labach,
  Jesse~C. Cresswell, Keyvan Golestan, Guangwei Yu, Anthony~L. Caterini, and
  Maksims Volkovs.
\newblock {TabDPT}: Scaling tabular foundation models on real data.
\newblock In \emph{Advances in Neural Information Processing Systems
  (NeurIPS)}, 2025.

\bibitem[M\"{u}ller et~al.(2022)M\"{u}ller, Hollmann, {Pineda Arango},
  Grabocka, and Hutter]{muller2021transformers}
Samuel M\"{u}ller, Noah Hollmann, Sebastian {Pineda Arango}, Josif Grabocka,
  and Frank Hutter.
\newblock Transformers can do bayesian inference.
\newblock In \emph{International Conference on Learning Representations
  (ICLR)}, 2022.

\bibitem[Nam et~al.(2023)Nam, Tack, Lee, Lee, and Shin]{nam2023stunt}
Jaehyun Nam, Jihoon Tack, Kyungmin Lee, Hankook Lee, and Jinwoo Shin.
\newblock {STUNT}: Few-shot tabular learning with self-generated tasks from
  unlabeled tables.
\newblock \emph{arXiv preprint arXiv:2303.00918}, 2023.

\bibitem[Qu et~al.(2025)Qu, Holzm\"{u}ller, Varoquaux, and {Le
  Morvan}]{qu2025tabicl}
Jingang Qu, David Holzm\"{u}ller, Ga\"{e}l Varoquaux, and Marine {Le Morvan}.
\newblock {TabICL}: A tabular foundation model for in-context learning on large
  data.
\newblock In \emph{Proceedings of the 42nd International Conference on Machine
  Learning (ICML)}. PMLR, 2025.

\bibitem[Somepalli et~al.(2021)Somepalli, Goldblum, Schwarzschild, Bruss, and
  Goldstein]{somepalli2021saint}
Gowthami Somepalli, Micah Goldblum, Avi Schwarzschild, C.~Bayan Bruss, and Tom
  Goldstein.
\newblock {SAINT}: Improved neural networks for tabular data via row attention
  and contrastive pre-training, 2021.

\bibitem[Thilak et~al.(2023)Thilak, Huang, Saremi, Dinh, Goh, Nakkiran,
  Susskind, and Littwin]{thilak2023lidar}
Vimal Thilak, Chen Huang, Omid Saremi, Laurent Dinh, Hanlin Goh, Preetum
  Nakkiran, Joshua~M. Susskind, and Etai Littwin.
\newblock {LiDAR}: Sensing linear probing performance in joint embedding {SSL}
  architectures, 2023.

\bibitem[Thimonier et~al.(2025)Thimonier, {De Melo Costa}, Popineau, Rimmel,
  and Doan]{thimonier2024tjepa}
Hugo Thimonier, Jos\'{e}~Lucas {De Melo Costa}, Fabrice Popineau, Arpad Rimmel,
  and Bich-Li\^{e}n Doan.
\newblock {T-JEPA}: Augmentation-free self-supervised learning for tabular
  data.
\newblock In \emph{International Conference on Learning Representations
  (ICLR)}, 2025.

\bibitem[Ucar et~al.(2021)Ucar, Hajiramezanali, and Edwards]{ucar2021subtab}
Talip Ucar, Ehsan Hajiramezanali, and Lindsay Edwards.
\newblock {SubTab}: Subsetting features of tabular data for self-supervised
  representation learning.
\newblock In \emph{Advances in Neural Information Processing Systems 34
  (NeurIPS)}, 2021.

\bibitem[Verdenius et~al.(2024)Verdenius, Zerio, and Wang]{verdenius2024latpfn}
Stijn Verdenius, Andrea Zerio, and Roy~L.M. Wang.
\newblock {LaT-PFN}: A joint embedding predictive architecture for in-context
  time-series forecasting, 2024.

\bibitem[Wang \& Sun(2022)Wang and Sun]{wang2022transtab}
Zifeng Wang and Jimeng Sun.
\newblock {TransTab}: Learning transferable tabular transformers across tables.
\newblock In \emph{Advances in Neural Information Processing Systems
  (NeurIPS)}, 2022.

\bibitem[Yoon et~al.(2020)Yoon, Zhang, Jordon, and van~der
  Schaar]{yoon2020vime}
Jinsung Yoon, Yao Zhang, James Jordon, and Mihaela van~der Schaar.
\newblock {VIME}: Extending the success of self- and semi-supervised learning
  to tabular domain.
\newblock In \emph{Advances in Neural Information Processing Systems 33
  (NeurIPS)}, 2020.

\bibitem[Zbontar et~al.(2021)Zbontar, Jing, Misra, LeCun, and
  Deny]{zbontar2021barlow}
Jure Zbontar, Li~Jing, Ishan Misra, Yann LeCun, and St\'{e}phane Deny.
\newblock {Barlow Twins}: Self-supervised learning via redundancy reduction.
\newblock In \emph{Proceedings of the 38th International Conference on Machine
  Learning (ICML)}. PMLR, 2021.

\bibitem[Zhang et~al.(2025)Zhang, Maddix, Yin, Erickson, Ansari, Han, Zhang,
  Akoglu, Faloutsos, Mahoney, Hu, Rangwala, Karypis, and Wang]{zhang2025mitra}
Xiyuan Zhang, Danielle~C. Maddix, Junming Yin, Nick Erickson, Abdul~Fatir
  Ansari, Boran Han, Shuai Zhang, Leman Akoglu, Christos Faloutsos, Michael~W.
  Mahoney, Cuixiong Hu, Huzefa Rangwala, George Karypis, and Bernie Wang.
\newblock {Mitra}: Mixed synthetic priors for enhancing tabular foundation
  models.
\newblock In \emph{Advances in Neural Information Processing Systems
  (NeurIPS)}, 2025.

\bibitem[Zhu et~al.(2023)Zhu, Shi, Erickson, Li, Karypis, and
  Shoaran]{zhu2023xtab}
Bingzhao Zhu, Xingjian Shi, Nick Erickson, Mu~Li, George Karypis, and Mahsa
  Shoaran.
\newblock {XTab}: Cross-table pretraining for tabular transformers.
\newblock In \emph{Proceedings of the 40th International Conference on Machine
  Learning (ICML)}. PMLR, 2023.

\end{thebibliography}
